\documentclass[a4paper,conference]{IEEEtran}
\usepackage{hyperref}
\usepackage{graphicx}
\usepackage{amsmath}

\usepackage{flushend}

\ifCLASSINFOpdf
\else
\fi
\usepackage{pgfplots}
\usepackage{tikz}
\pgfplotsset{compat=1.16}

\usepackage{booktabs}
\usepackage{subcaption}

\usetikzlibrary{shapes,arrows,calc,positioning,matrix,fit,backgrounds}

\tikzstyle{conv} = [rectangle, draw, fill=blue!20, text width=10em, text centered, rounded corners, minimum height=2em]
\tikzstyle{norm} = [rectangle, draw, fill=purple!20, text width=10em, text centered, rounded corners, minimum height=2em]
\tikzstyle{activation} = [rectangle, draw, fill=green!20, text width=10em, text centered, rounded corners, minimum height=2em]
\tikzstyle{layer} = [rectangle, draw, fill=yellow!20, text width=10em, text centered, rounded corners, minimum height=2em]
\tikzstyle{module} = [rectangle, draw, fill=orange!20, text width=10em, text centered, rounded corners, minimum height=2em]
\tikzstyle{modules} = [rectangle, draw, fill=red!20, text width=10em, text centered, rounded corners, minimum height=2em]

\tikzstyle{op} = [circle, draw, text centered, minimum height=1em]
\tikzstyle{split} = [circle, fill=black, inner sep=0pt, minimum size=0.5em]

\tikzstyle{line} = [draw, -latex']

\newcommand{\lone}{\ell^1}
\newcommand{\ltwo}{\ell^2}

\newcommand*\ratio[1]{\cleanratio#1\relax}
\def\cleanratio#1:#2\relax{#1\mathord{:}#2}

\newcommand{\multidim}[1]{\mathbf{#1}}

\begin{document}
%
\title{Object Tracking by Detection with Visual and Motion Cues}

\author{\IEEEauthorblockN{Niels Ole Salscheider}
\IEEEauthorblockA{FZI Research Center for Information Technology\\
Haid-und-Neu-Str. 10-14\\
76131 Karlsruhe\\
Germany\\
Email: salscheider@fzi.de}

}


\maketitle

\begin{abstract}
Self-driving cars and other autonomous vehicles need to detect and track objects in camera images.
We present a simple online tracking algorithm that is based on a constant velocity motion model with a Kalman filter, and an assignment heuristic.
The assignment heuristic relies on four metrics:
An embedding vector that describes the appearance of objects and can be used to re-identify them, a displacement vector that describes the object movement between two consecutive video frames, the Mahalanobis distance between the Kalman filter states and the new detections, and a class distance.
These metrics are combined with a linear SVM, and then the assignment problem is solved by the Hungarian algorithm.
We also propose an efficient CNN architecture that estimates these metrics.
Our multi-frame model accepts two consecutive video frames which are processed individually in the backbone, and then optical flow is estimated on the resulting feature maps.
This allows the network heads to estimate the displacement vectors.
We evaluate our approach on the challenging BDD100K tracking dataset.
Our multi-frame model achieves a good MOTA value of 39.1\,\% with low localization error of 0.206 in MOTP.
Our fast single-frame model achieves an even lower localization error of 0.202 in MOTP, and a MOTA value of 36.8\,\%.
\end{abstract}


%
\IEEEpeerreviewmaketitle

\section{Introduction}
Self-driving cars, other autonomous vehicles, and advanced driver assistance systems need an accurate perception of the environment.
This information is crucial to make informed decisions.
Popular sensors include cameras, radar and lidar.
Amongst these, camera images contain the most detailed information, but they are also hard to interpret.
Efficient Convolutional Neural Networks (CNNs) have emerged in recent years that are up to the challenge.
They show strong performance in the relevant tasks of object detection and pixel-wise semantic segmentation.

But object detection in single images is not enough for the aforementioned applications.
A detection from a single image neither contains information about the velocity of the corresponding object, nor about its past trajectory and behaviour.
This information is however important to predict its future behaviour, and to plan accordingly.

This problem is usually solved by tracking each object over time.
In the past, detection-free approaches like the correlation filter \cite{Bolme2010,Henriques2015,Danelljan2015} or siamese networks \cite{Bertinetto2016,Dong2018,He2018,Li2018} were popular.
These algorithms are initialized once with a detection, and then track the corresponding image patch based on visual features.

With the emergence of highly accurate object detectors, tracking-by-detection approaches became more popular.
These approaches assume that the detector has a low false negative rate and therefore detects nearly all objects.
The tracking approach then associates detections to existing tracks, and manages the lifetime of tracks.
Most approaches also rely on a motion model.
It filters noise in the object localization and allows to predict the object state for time steps without a corresponding detection.

Tracking approaches can either be online or offline.
Offline approaches do not have to meet real-time requirements.
They also have access to the whole video sequence which allows acausal reasoning.
Graph-based methods that optimize over the whole video sequence are popular and can achieve high accuracies \cite{Zhang2008,Pirsiavash2011,Butt2013}.
Offline methods are however not suitable for the use in autonomous vehicles.
This application requires online methods that use little computational resources and only perform causal computations.
Simple online methods often only perform frame-by-frame associations.

Many object detection algorithms are oblivious to the tracking task and only output detections.
In this work, we however present a detection approach that provides additional information to aid the association and tracking task.
This additional information includes appearance feature vectors to re-identify objects, and displacement vectors that describe the delta movements of objects between frames.
We estimate the displacement vectors by passing two consecutive frames to our CNN backbone and correlating the resulting feature maps.
We then solve the association problem with a Support Vector Machine (SVM) and the Hungarian algorithm.
Finally, the objects are tracked using a constant velocity motion model with a Kalman filter \cite{Kalman1960}.
We learn the parameters for the SVM and the Kalman filter from labelled data.

The remainder of this paper is structured as follows:
We present related work in Section~\ref{sec:related_work}.
We then describe our approach in Section~\ref{sec:approach}, and evaluate it in Section~\ref{sec:evaluation}.
Section~\ref{sec:conclusion} finally concludes this paper.

\subsection{Related Work}
\label{sec:related_work}

An early work \cite{Bewley2016} shows that a simple tracking approach can achieve state of the art performance when used together with a strong object detector.
The authors use a constant velocity motion model with a Kalman filter.
When a new camera frame is recorded, the prediction step of the Kalman filter is executed.
This yields estimates for the current locations of objects.
Then, the new detections are associated to the existing tracks only based on the pairwise Intersection over Union (IoU) values.
Afterwards, the Kalman filter is updated with the new measurements.

\cite{Wojke2017} proposes an extension to this approach.
It additionally relies on visual appearance information in the association step.
This reduces the number of identity switches by 45\,\%.

In \cite{Feichtenhofer2017}, the authors propose a tracking approach that is similar to ours.
It also passes two consecutive frames through a CNN backbone and correlates the feature maps.
The results of this are used to detect bounding boxes, and to estimate the displacement of the bounding boxes between both frames.
Complete tracks are then recovered by solving a maximisation problem with the Viterbi algorithm.

Traktor is a tracking-by-detection approach which also has some similarities with ours \cite{Bergmann2019}.
It exploits the regression head of a two-stage object detector for tracking:
The RoI-pooled backbone features of the current frame, based on the previous object bounding box, are passed into the regression head.
The regression head  then regresses the box parameters of the new object location.
The authors also present two extensions to their simple tracker:
The first extension is to use a Siamese network for the re-identification task.
The second extension is to use a motion model, and to compensate for the camera movement.

In \cite{Luo2018}, the authors present Fast and Furious.
Their approach performs object detection, tracking and motion forecasting in birds-eye-view (BEV) images.
The authors generate ego-motion compensated BEV images, and then concatenate BEV images of multiple time steps to incorporate temporal information.
This information is fed into a single-stage object detector that directly regresses bounding boxes, both for the current time step and for predicted future locations.
Tracks are formed by associating current detections to the predictions made in past time steps.
The association is performed based on the distance of the corresponding bounding boxes.

\section{Approach}
\label{sec:approach}
We propose an object detection and tracking method that is simple, fast, and accurate.
It employs a tracking-by-detection approach based on a Kalman filter, and a track management and association heuristic.
The detections, as well as additional features for track association, are generated by our CNN architecture which comes in a single-frame and a multi-frame variant.

The single-frame variant takes a single input image, while the multi-frame variant takes pairs of consecutive images.
Each object detection consists of an objectness score, a class score, regressed bounding box parameters, and an embedding vector.
This embedding vector describes the appearance of the object and can be used as feature for re-identification.
The multi-frame variant additionally predicts a displacement vector for each detection.
This displacement vector allows to calculate the position of a detected object in the previous frame, given the position in the current frame.
Section~\ref{sec:cnn_structure} describes our CNN architecture in more detail.

Our tracking approach is based on a simple heuristic for track creation and deletion, and uses a constant velocity motion model with a Kalman filter.
We calculate pairwise assignment costs between existing tracks and new detections, and employ the Hungarian algorithm to solve the assignment problem.
These pairwise costs are predicted by an SVM based on multiple features.
They depend on the predicted object class, the Mahalanobis distance between the detection bounding box and the Kalman filter state, and the embedding and displacement vectors from the CNN.
More details are provided in Section~\ref{sec:tracking}.

\subsection{CNN Architecture}
\label{sec:cnn_structure}

Our tracking approach can be used with a single-frame and a multi-frame CNN architecture.
A multi-frame architecture allows to also predict displacement vectors for the detected objects.
This is because it has access to two consecutive images of the video stream.
The displacement vectors can be used as an additional hint in the association heuristic.
In the following, we first describe the architecture of the single-frame model.
We then describe our proposed modifications for the multi-frame model.

\subsubsection{Single-Frame Model}

Our single-frame model is a slightly modified EfficientDet-D3 \cite{Tan2020}.
The EfficientNet \cite{Tan2019} backbone generates feature maps at different resolution levels by successive down-sampling.
These feature maps are then fed into a BiFPN to fuse information at different feature levels.
BiFPN blocks were proposed by the authors of EfficientDet.
It takes ideas from FPN \cite{Lin2017_2}, PANet \cite{Liu2018} and NAS-FPN \cite{Ghiasi2019} and improves upon them.

In the following, $P_i$ refers to the level with feature map resolutions of $1/2^i$ of the input image resolution.
The original EfficientNet architecture feeds the last feature maps produced by the backbone at the levels $P_3$ to $P_7$ to the BiFPN.
We, however, also include feature level $P_2$ in the BiFPN.
One reason for this is that we also want to detect very small objects, like distant traffic lights.
Another reason is that the multi-frame model correlates feature maps inside the BiFPN, as will be explained shortly.
By including the $P_2$ feature level, the correlation layer has access to a feature map with higher resolution.
This can lead to more accurate displacement estimates.

Like in the original EfficientNet architecture, the output feature maps of the BiFPN are passed to RetinaNet-style \cite{Lin2017} object detection heads.
EfficientNet has one head to predict the presence of objects of certain classes for all anchor boxes, and one head to regress the corresponding bounding box parameters.
We replace the first head with two separate ones in our architecture:
We employ one head that predicts an objectness score, and one that predicts class scores for all classes of interest.
In other words, we predict the probability that there is an object $\Pr(\text{object})$, and the conditional class probability $\Pr(\text{class} = c_i \vert \text{object})$, while EfficientDet only predicts the joint probability $\Pr(\text{object}, \text{class} = c_i)$.
This allows us to detect objects even when we are unsure about the class (i.\,e. when $\Pr(\text{class} = c_i \vert \text{object}) < 0.5 \quad \forall c_i$).
This is important in the context of self-driving cars where collisions with any kind of object must be avoided.
We train both of these heads with Focal Loss \cite{Lin2017}.

We keep the network head for bounding box regression unchanged, and train it with smooth $\lone$ loss.
We however change the set of aspect ratios of the corresponding anchor boxes to $\{\ratio{1:4}, \ratio{1:2}, \ratio{1:1}, \ratio{2:1}, \ratio{4:1}\}$.
We add the aspect ratios of $\ratio{1:4}$ and $\ratio{4:1}$ to detect more asymmetric objects like trains or pedestrians.

We also add another network head that predicts a feature vector consisting of 32 floating point numbers for each detection.
This feature vector is trained with a weighted version of Margin Loss \cite{Wu2017}:
Feature vectors of anchor boxes that are assigned to the same ground truth object are trained to have an $\ltwo$ distance below a threshold $\beta$ with margin $\alpha$.
Feature vectors of anchor boxes that correspond to different objects, on the other hand, are trained to have an $\ltwo$ distance above the threshold $\beta$ with margin $\alpha$.
We use this feature vector to perform non-maximum suppression as in \cite{Salscheider2020_2}.
We also use it as an appearance hint in our heuristic to assign detections to existing tracks.

We weight the losses of the different training objectives according to \cite{Kendall2017}.

\subsubsection{Multi-Frame Model}

Our multi-frame architecture is based on the single-frame architecture.
Each image is processed by the backbone and the first half of the BiFPN individually.
After this, a small sub-network is inserted.
The task of this sub-network is to calculate the flow field between the two consecutive frames based on the BiFPN feature maps.
Then, the feature map of the current image, the feature map of the last image warped by the flow field, and the flow field itself are concatenated.
The number of channels is reduced by a $1 \times 1$ convolution.
The result is fed into the second half of the BiFPN and then into the network heads.
This is visualized in Figure~\ref{figure:multi_frame_arch} and Figure~\ref{figure:warping_module}.

\begin{figure}
 \begin{center}
   \begin{tikzpicture}
    \footnotesize

    \node (input) [color=black!50] at (0, 0) {$\multidim{I}_1$};

    \node[modules, color=black!50, fill=red!7, below of=input, text width=8em] (m1) {};
    \node[modules, color=black!50, fill=red!7, below of=m1, text width=7em] (m2) {};
    \node[modules, color=black!50, fill=red!7, below of=m2, text width=6em] (m3) {};
    \node[modules, color=black!50, fill=red!7, below of=m3, text width=5em] (m4) {};
    \node[modules, color=black!50, fill=red!7, below of=m4, text width=4em] (m5) {};
    \node[modules, color=black!50, fill=red!7, below of=m5, text width=3em] (m6) {};
    \node[modules, color=black!50, fill=red!7, below of=m6, text width=2em] (m7) {};
    \node[modules, color=black!50, fill=red!7, below of=m7, text width=1em] (m8) {};

    \node[module, color=black!50, fill=orange!10, text width=20em,rotate=90] at ($0.5*(m3)+0.5*(m8)+(2.3,0)$) (fpn) {};

    \path [line,color=black!50] (input) -- (m1);
    \path [line,color=black!50] (m1) -- (m2);
    \path [line,color=black!50] (m2) -- (m3);
    \path [line,color=black!50] (m3) -- (m4);
    \path [line,color=black!50] (m4) -- (m5);
    \path [line,color=black!50] (m5) -- (m6);
    \path [line,color=black!50] (m6) -- (m7);
    \path [line,color=black!50] (m7) -- (m8);

    \path [line,color=black!50] (m3) -- (m3 -| fpn.north);
    \path [line,color=black!50] (m4) -- (m4 -| fpn.north);
    \path [line,color=black!50] (m5) -- (m5 -| fpn.north);
    \path [line,color=black!50] (m6) -- (m6 -| fpn.north);
    \path [line,color=black!50] (m7) -- (m7 -| fpn.north);
    \path [line,color=black!50] (m8) -- (m8 -| fpn.north);

    \node (input_2) at (-0.3, -0.3) {$\multidim{I}_2$};

    \node[modules, below of=input_2, text width=8em] (m1_2) {};
    \node[modules, below of=m1_2, text width=7em] (m2_2) {};
    \node[modules, below of=m2_2, text width=6em] (m3_2) {};
    \node[modules, below of=m3_2, text width=5em] (m4_2) {};
    \node[modules, below of=m4_2, text width=4em] (m5_2) {};
    \node[modules, below of=m5_2, text width=3em] (m6_2) {};
    \node[modules, below of=m6_2, text width=2em] (m7_2) {};
    \node[modules, below of=m7_2, text width=1em] (m8_2) {};

    \draw [thick,dotted] let \p1 = (m1.north east) in let \p2 = (m1_2.north west) in let \p3=(m8_2.south) in ($(\x1, \y1)+(0.2,0.2)$) rectangle ($(\x2, \y3)+(-0.2,-0.2)$);
    \node [align=left,below=0.4 of m8_2, text width=12em] {EfficientNet backbones};

    \node[module, text width=20em,rotate=90] at ($0.5*(m3_2)+0.5*(m8_2)+(2.3,0)$) (fpn_2) {BiFPN (repeated blocks)};

    \path [line] (input_2) -- (m1_2);
    \path [line] (m1_2) -- (m2_2);
    \path [line] (m2_2) -- (m3_2);
    \path [line] (m3_2) -- (m4_2);
    \path [line] (m4_2) -- (m5_2);
    \path [line] (m5_2) -- (m6_2);
    \path [line] (m6_2) -- (m7_2);
    \path [line] (m7_2) -- (m8_2);

    \path [line] (m3_2) -- (m3_2 -| fpn_2.north);
    \path [line] (m4_2) -- (m4_2 -| fpn_2.north);
    \path [line] (m5_2) -- (m5_2 -| fpn_2.north);
    \path [line] (m6_2) -- (m6_2 -| fpn_2.north);
    \path [line] (m7_2) -- (m7_2 -| fpn_2.north);
    \path [line] (m8_2) -- (m8_2 -| fpn_2.north);

    \node[modules, text width=20em,rotate=90] at ($0.5*(fpn)+0.5*(fpn_2)+(1.0,0)$) (flow) {Flow estimation \& warping};
    \node[module, text width=20em,rotate=90,below=0 of flow] (fpn3) {BiFPN (repeated blocks)};

    \path [line,color=black!50] (m3 -| fpn.south) -- (m3 -|flow.north);
    \path [line,color=black!50] (m4 -| fpn.south) -- (m4 -|flow.north);
    \path [line,color=black!50] (m5 -| fpn.south) -- (m5 -|flow.north);
    \path [line,color=black!50] (m6 -| fpn.south) -- (m6 -|flow.north);
    \path [line,color=black!50] (m7 -| fpn.south) -- (m7 -|flow.north);
    \path [line,color=black!50] (m8 -| fpn.south) -- (m8 -|flow.north);

    \path [line] (m3_2 -| fpn_2.south) -- (m3_2 -|flow.north);
    \path [line] (m4_2 -| fpn_2.south) -- (m4_2 -|flow.north);
    \path [line] (m5_2 -| fpn_2.south) -- (m5_2 -|flow.north);
    \path [line] (m6_2 -| fpn_2.south) -- (m6_2 -|flow.north);
    \path [line] (m7_2 -| fpn_2.south) -- (m7_2 -|flow.north);
    \path [line] (m8_2 -| fpn_2.south) -- (m8_2 -|flow.north);

    \node at ($0.5*(m3)+0.5*(m3_2)$) (help1) {};
    \node[modules, text width=4em, right=4.8 of help1] (h3) {Heads};
    \node[modules, text width=4em, below of=h3] (h4) {Heads};
    \node[modules, text width=4em, below of=h4] (h5) {Heads};
    \node[modules, text width=4em, below of=h5] (h6) {Heads};
    \node[modules, text width=4em, below of=h6] (h7) {Heads};
    \node[modules, text width=4em, below of=h7] (h8) {Heads};

    \node[right=0.3 of h3] (o3) {};
    \node[right=0.3 of h4] (o4) {};
    \node[right=0.3 of h5] (o5) {};
    \node[right=0.3 of h6] (o6) {};
    \node[right=0.3 of h7] (o7) {};
    \node[right=0.3 of h8] (o8) {};

    \path [line] (fpn3.south |- h3) -- (h3) node[pos=0.5, above] (desc3a) {$P_2$};
    \path [line] (fpn3.south |- h4) -- (h4);
    \node [below of=desc3a] (desc4a) {$P_3$};
    \path [line] (fpn3.south |- h5) -- (h5);
    \node [below of=desc4a] (desc5a) {$P_4$};
    \path [line] (fpn3.south |- h6) -- (h6);
    \node [below of=desc5a] (desc6a) {$P_5$};
    \path [line] (fpn3.south |- h7) -- (h7);
    \node [below of=desc6a] (desc7a) {$P_6$};
    \path [line] (fpn3.south |- h8) -- (h8);
    \node [below of=desc7a] (desc8a) {$P_7$};

    \path [line] (h3) -- (o3);
    \path [line] (h4) -- (o4);
    \path [line] (h5) -- (o5);
    \path [line] (h6) -- (o6);
    \path [line] (h7) -- (o7);
    \path [line] (h8) -- (o8);
  \end{tikzpicture}
 \end{center}
 \caption{Our proposed multi-frame architecture.
 The input images are first processed individually by the backbone and the first half of the BiFPN.
 Then, optical flow is estimated, and the feature maps of the frame from the previous time step are warped by the flow estimate.
 This aligns them with the feature maps of the current frame.
 The flow estimation and warping module is visualized in Figure~\ref{figure:warping_module}.
 Finally, the output is processed by the network heads.}
  \label{figure:multi_frame_arch}
\end{figure}
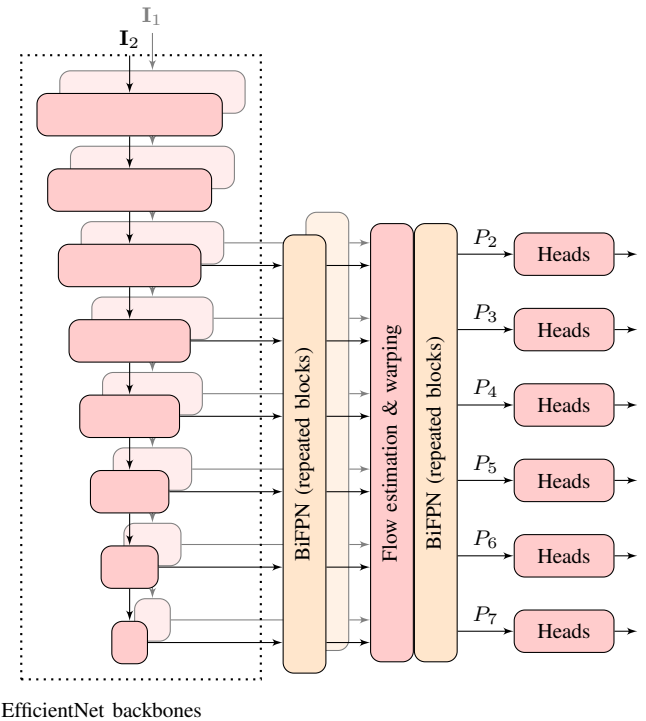

\begin{figure}
 \begin{center}
  \begin{tikzpicture}
    \footnotesize
    \node [text width=3em, text centered] (input_f1) {$\multidim{f}_{\text{i1}, l}$};
    \node [text width=3em, text centered, right of=input_f1] (input_f2) {$\multidim{f}_{\text{i2}, l}$};
    \node [text width=3em, text centered, right of=input_f2] (input_flow) {$\multidim{o}_{l + 1}$};
    \node [text width=3em, text centered, right of=input_flow] (input_mask) {$\multidim{m}_{l + 1}$};
    \node [text width=3em, text centered, right of=input_mask] (input_flowfeat) {$\multidim{f}_{\text{sn}, l+1}$};

    \node[split,below=0.3 of input_f1] (split1) {};
    \node[split,below=0.6 of input_f2] (split2) {};

    \node[module, text width=17em, below=1.1 of input_flow] (flow_est) {Flow estimation};
    \node[layer, text width=17em, below=1.4 of flow_est] (warp) {warp};
    \node[op, below=0.3 of warp] (concat) {$\oplus$};
    \node[conv, text width=17em, below=0.3 of concat] (conv) {Conv ($1 \times 1$)};
    \node[below=0.3 of conv] (out) {$\multidim{f}_{\text{fused}, l}$};

    \node[split] (split3) at ($0.25*(flow_est.south)+0.75*(warp.north)$) {};
    \node (help6) at ($0.5*(flow_est.south)+0.5*(warp.north)$) {};

    \coordinate (help2) at ($(flow_est.west)+(-0.6,0)$);
    \coordinate (help3) at ($(flow_est.west)+(-0.3,0)$);
    \coordinate (help4) at ($0.5*(input_f1)+0.5*(input_f2)$);
    \coordinate (help5) at ($0.25*(flow_est.south east)+0.75*(warp.north east)$);
    \coordinate (help8) at ($0.75*(flow_est.south east)+0.25*(warp.north east)$);
    \coordinate (help8a) at ($0.5*(flow_est.south east)+0.5*(warp.north east)$);

    \node (help7) at ($(help8)+(0.3,0)$) {};
    \node (help7a) at ($(help8a)+(0.3,0)$) {};
    \node[split] (split4) at ($(help5)+(0.3,0)$) {};
    \node [text width=3em, right of=split4] (out_flow) {$\multidim{o}_{l}$};
    \node [text width=3em, right of=help7a] (out_mask) {$\multidim{m}_{l}$};
    \node [text width=3em, right of=help7] (out_feat) {$\multidim{f}_{\text{sn}, l}$};

    \path [line] (input_flow) -- (input_flow|-flow_est.north);
    \path [line] (input_mask) -- (input_mask|-flow_est.north);
    \path [line] (input_flowfeat) -- (input_flowfeat|-flow_est.north);
    \path [line] (input_f1) -- (split1);
    \path [line] (input_f2) -- (split2);
    \path [line] (split1) -- (split1|-flow_est.north);
    \path [line] (split2) -- (split2|-flow_est.north);
    \path [line] (flow_est) -- (split3);
    \path [line] (split3) -- (warp);
    \path [line] (warp) -- (concat);
    \path [line] (concat) -- (conv);
    \path [line] (conv) -- (out);
    \path [line] (split1) -| (help2) |- (concat);
    \path [line] (split3) -- (split4);
    \path [line] (split4) |- (concat);
    \path [line] (split4) -- (out_flow);
    \path [line] (flow_est.south-|input_mask) |- (help8a) -- (out_mask);
    \path [line] (flow_est.south-|input_flowfeat) |- (help8) -- (out_feat);
    
    \path [line] let \p1=(help6) in let \p2=(help4) in  (split2) -| (help3) |- (\x2, \y1) -- (\x2, \y1|-warp.north);
  \end{tikzpicture}
 \end{center}
 \caption{Our proposed warping module that aligns the feature maps of two consecutive frames.
 It is instantiated for each feature level $l$ of the BiFPN.
 The inputs $\multidim{f}_{\text{i1}, l}$ and $\multidim{f}_{\text{i2}, l}$ are the output feature maps of the BiFPN for the first and second image, respectively.
 The inputs $\multidim{o}_{l + 1}$, $\multidim{m}_{l + 1}$ and $\multidim{f}_{\text{sn}, l+1}$ are the flow estimate, a corresponding mask and intermediate feature maps from feature level $l + 1$ (if available).
 They are produced and consumed by the flow estimation module which is visualized in Figure~\ref{figure:flow_module}.
 Each module produces a feature map $\multidim{f}_{\text{fused}, l}$ that is fed to the second half of the BiFPN.
 }
  \label{figure:warping_module}
\end{figure}
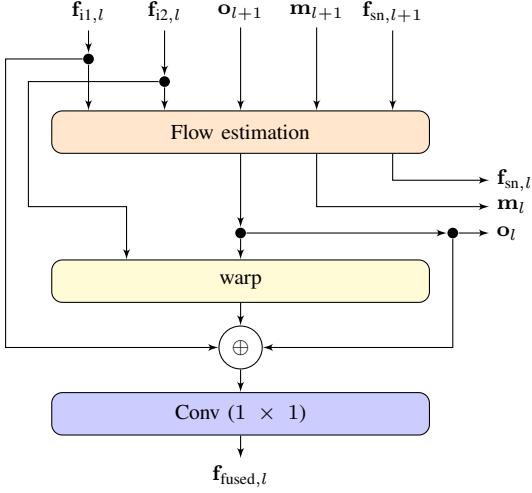

\begin{figure}
 \begin{center}
  \begin{tikzpicture}
    \footnotesize
    \node [text width=3em] (input_flowfeat) {$\multidim{f}_{\text{sn}, l+1}$};
    \node [text width=3em, below=0.4 of input_flowfeat] (input_mask) {$\multidim{m}_{l + 1}$};
    \node [text width=3em, below=0.4 of input_mask] (input_flow) {$\multidim{o}_{l + 1}$};
    \node [text width=3em, below=0.8 of input_flow] (input_f2) {$\multidim{f}_{\text{i2}, l}$};
    \node [text width=3em, below=1.3 of input_f2] (input_f1) {$\multidim{f}_{\text{i1}, l}$};

    \node [layer, text width=6em, right=0.6 of input_flowfeat] (upsample_flowfeat) {upsample};
    \node [layer, text width=6em, right=0.6 of input_mask] (upsample_mask) {upsample};
    \node [layer, text width=6em, right=0.6 of input_flow] (upsample_flow) {upsample};
    \node [layer, text width=6em, right=0.6 of input_f2] (warp) {warp};
    \node [layer, text width=6em, right=0.6 of input_f1] (correlate) {correlate};

    \node [op, above=0.5 of correlate] (mul_mask) {$\cdot$};

    \node [split] (split_f1) at ($0.5*(input_f1.east)+0.5*(correlate.west)$) {};
    \node [split] (split_up) at ($0.5*(upsample_flow.south)+0.5*(warp.north)$) {};
    \coordinate (help1) at ($(correlate.south)+(0.0,-0.2)$);
    \coordinate (help2) at ($(upsample_mask.east)+(0.2,0.0)$);

    \node [op, right=1.6 of split_up] (concat) {$\oplus$};
    \node [module, text width=6em, right=0.4 of concat] (dnet) {CNN};
    \node [split, below=0.3 of dnet] (split_output) {};
    \node [conv, text width=6em, below=0.3 of split_output] (of_est) {Conv ($3 \times 3$)};
    \node [conv, text width=6em, below=0.3 of of_est] (mask_est) {Conv ($3 \times 3$)};
    \node [activation, text width=6em, below=0 of mask_est] (mask_sig) {sigmoid};

    \coordinate (help3) at ($(mask_est.west)+(-0.2,0.0)$);

    \node [text width=3em, right=0.4 of dnet] (out_flowfeat) {$\multidim{f}_{\text{sn}, l}$};
    \node [text width=3em, right=0.4 of of_est] (out_flow) {$\multidim{o}_{l}$};
    \node [text width=3em, right=0.4 of mask_sig] (out_mask) {$\multidim{m}_{l}$};
    
    \path [draw,thick,dotted] let \p1 = (mul_mask.south) in let \p2 = (split_f1) in let \p3 = (split_up) in let \p4 = (input_flowfeat.west) in let \p5 = (input_flowfeat.north) in let \p6 = (upsample_flowfeat.north east) in let \p7 = (help2) in ($(\x4, \y6)+(-0.2,0.2)$) -- ($(\x7, \y6)+(0.2,0.2)$) -- ($(\x7, \y1)+(0.2,-0.2)$) -- ($(\x2, \y1)+(0.0,-0.2)$) -- ($(\x2, \y3)$) -- ($(\x4, \y3)+(-0.2,0.0)$) --  ($(\x4, \y6)+(-0.2,0.2)$);
    \node [align=left] at ($(input_flowfeat.north)+(1.2,0.5)$) {For all levels but the highest};

    \path [line] (input_flowfeat) -- (upsample_flowfeat);
    \path [line] (input_mask) -- (upsample_mask);
    \path [line] (input_flow) -- (upsample_flow);
    \path [line] (input_f2) -- (warp);
    \path [line] (input_f1) -- (split_f1);
    \path [line] (split_f1) -- (correlate);
    \path [line] (input_f2) -- (warp);
    \path [line] (split_up) -- (warp);
    \path [line] (upsample_flow) -- (split_up);
    \path [line] (warp) -- (mul_mask);
    \path [line] (upsample_mask) -- (help2) |- (mul_mask);
    \path [line] (mul_mask) -- (correlate);

    \path [line] (split_up) -- (concat);
    \path [line] (upsample_flowfeat) -| (concat);
    \path [line] (correlate) edge [out=0, in=250] (concat);
    \path [line] (split_f1) |- (help1) -| (concat);
    \path [line] (concat) -- (dnet);
    \path [line] (dnet) -- (out_flowfeat);
    \path [line] (of_est) -- (out_flow);
    \path [line] (mask_sig) -- (out_mask);

    \path [line] (dnet) -- (split_output);
    \path [line] (split_output) -- (of_est);
    \path [line] (split_output) -| (help3) |- (mask_est);

  \end{tikzpicture}
 \end{center}
 \caption{Our flow estimation module which is based on PWC-Net \cite{Sun2018}.
 The blocks inside the dotted line are omitted for the highest feature level, because no optical flow estimate $\multidim{o}_{l + 1}$ exists for it.
 In contrast to PWC-Net, our CNN block does not consist of convolutions with DenseNet \cite{Huang2017} connections.
 Instead, it has the same structure as the EfficientDet network heads.
 We also predict a mask $\multidim{m}_{l}$ at each feature level $l$.
 It is used to mask the pixels with unobservable optical flow.
 }
 \label{figure:flow_module}
\end{figure}
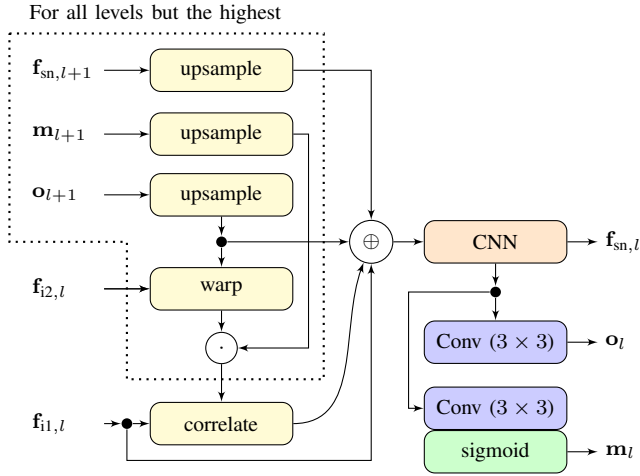

Our optical flow estimation module is based on PWC-Net \cite{Sun2018}.
It is visualized in Figure~\ref{figure:flow_module}.
Feature levels $P_2$ to $P_7$ form a feature pyramid, so that only partial correlation cost volumes have to be computed---The flow predicted at a level with lower resolution limits the search space at the next level in the pyramid, and so on.
That means that the flow is calculated starting from level $P_7$ to level $P_2$, and at each level the estimate is refined.

At each level $l$, the feature map $\multidim{f}_{\text{i2}, l}$ of the previous frame is shifted by the flow estimate $\multidim{o}_{l + 1}$ of the previous level $l + 1$ (except for level $P_7$ where no estimate exists yet).
Then, the partial correlation cost volume is computed between the warped feature map of the previous frame and the feature map of current frame $\multidim{f}_{\text{i1}, l}$.
The resulting cost volume, both feature maps, and the upsampled flow of the previous level (if available) are concatenated.
A small sub-network then estimates the refined flow $\multidim{o}_l$ based on this input.
It also estimates a mask $\multidim{m}_l$ that indicates which pixels have unobservable optical flow.
The optical flow is unobservable for pixels for which the surrounding image patch is not visible in the previous frame.

In theory, this sub-network for flow estimation can be placed anywhere in the BiFPN.
The only requirement is that the feature levels $P_2$ to $P_7$ are available.
We chose the middle of the BiFPN, because here information at different scales has already been fused to some extent.
This can help to find robust correlations.
The second half of the BiFPN can then also fuse the information from the optical flow estimates at different scales.

The proposed design of the multi-frame architecture provides the network with information about the movement between frames.
This allows the network to estimate a displacement vector for each bounding box detection.
We add an additional network head to the multi-frame architecture for this task.
Each displacement vector indicates where the center of the corresponding bounding box was in the frame before, relative to the position in the current frame.

We train the outputs of the multi-frame model that also exist in the single-frame model as before.
We use smooth $\lone$ loss to train the output of the network head for the displacement estimation.
The optical flow estimates at each level are trained with a supervised and a self-supervised loss.
We upsample all flow estimates to match the original image size before calculating the loss.
Please note that also the magnitude of the flow estimate has to be scaled accordingly.

We use smooth $\lone$ loss for the flow estimates for the supervised loss.
In both cases, the mask outputs are trained with a regularizing term that pulls them towards one.
For the self-supervised loss, we use photometric loss terms to train the flow estimates.
These loss terms are the same that are used in \cite{Godard2019} to learn monocular depth estimation.

\subsection{Association and Tracking}
\label{sec:tracking}

Our proposed tracking approach is conceptionally simple and consists of a motion model and an assignment and track management heuristic.

\subsubsection{Motion Model}
We use a constant velocity motion model with a Kalman filter.
The state vector of the Kalman filter is $(p_x, p_y, p_w, p_h, v_x, v_y, v_w, v_h)^T$.
The components $p_x$ and $p_y$ are the center coordinates of the object bounding box, and $p_w$ and $p_h$ are the width and height.
The components $v_x$, $v_y$, $v_w$, and $v_h$ are the corresponding velocities.
With this, the state transition matrix is as follows:
\[
 \multidim{F}_k = \begin{pmatrix}
                 1 & 0 & 0 & 0 & dt & 0 & 0 & 0 \\
                 0 & 1 & 0 & 0 & 0 & dt & 0 & 0 \\  
                 0 & 0 & 1 & 0 & 0 & 0 & dt & 0 \\
                 0 & 0 & 0 & 1 & 0 & 0 & 0 & dt \\
                 0 & 0 & 0 & 0 & 1 & 0 & 0 & 0 \\
                 0 & 0 & 0 & 0 & 0 & 1 & 0 & 0 \\
                 0 & 0 & 0 & 0 & 0 & 0 & 1 & 0 \\
                 0 & 0 & 0 & 0 & 0 & 0 & 0 & 1
                \end{pmatrix}
\]
Here, $dt$ is the time between two consecutive frames.
The observation matrix $\multidim{H}_k$ is trivial since $p_x$, $p_y$, $p_w$ and $p_h$ are directly observed.
The multi-frame model can also directly observe $v_x$ and $v_y$---this information is provided by the displacement estimates.

We estimate the covariance matrices of the process noise $\multidim{Q}_k$ and of the observation noise $\multidim{R}_k$ from data.
This is done using gradient descent, as proposed in \cite{Abbeel2005}.
We generate the training data for this by running our CNN on a training dataset, and then manually assigning the detections to the ground truth tracks.

\subsubsection{Track Management and Association Heuristic}
In each time step, our CNN architecture generates new bounding box detections.
We first predict the states of all existing tracks to the current time step.
This is done by executing the prediction step of the corresponding Kalman filters.
We then associate the new detections to existing tracks.
Afterwards, we execute the update step of the Kalman filters of all tracks with associated detections.
This incorporate the new measurements.
There can however be tracks without assigned detections, or detections without assigned tracks.
We create new tracks for unassigned detections, but only mark them as stable after 3 consecutive observations.
Tracks without observations during the last 0.5\,s are deleted.

Our association heuristic uses multiple costs to decide if a detection belongs to an existing track, and if so, to which.
These costs are calculated for each pair of existing tracks (after being predicted to the current time step) and new detections:
\begin{itemize}
 \item A similarity metric based on the embedding vectors trained with Margin Loss.
  For each track, we store the embedding vectors of the last 10 detections that where assigned to the track.
  The $\ltwo$ distance between the embedding vector of the new detection and these stored embedding vectors is calculated.
  The final cost is then the minimum of the calculated $\ltwo$ distances.
 \item The $\ltwo$ distance between the bounding box position in the last time stamp, and the estimated old position based on the current detection.
  The latter is calculated by adding the estimated displacement vector to the position of the current detection.
  This information is only available when using the multi-frame model.
 \item The Mahalanobis distance between the new measurement and the state of the Kalman filter after the prediction step.
  This cost indicates how well the new measurement fits to the predicted state.
 \item A cost for the predicted class.
  We assign a cost of 0 if the predicted class of the new detection matches the current class of the track.
  Otherwise, we assign a cost of 1.
  It is possible to improve the heuristic by returning different costs, depending on how likely it is to mix up two given classes.
  We however refrain from this for simplicity.
\end{itemize}

Based on these metrics, we train a linear SVM to predict if a detection belongs to a given track or not.
This means that we only have to calculate a weighted sum of the metrics above and a bias term during inference.
The sign of the result indicates if the detection belongs to the given track or not.
We train the SVM with the same dataset that we use to estimate the parameters of the Kalman filter.

If the SVM classifier was perfect and the features would allow for a perfect prediction, then the association problem would already be solved.
In practice it can however happen that the SVM predicts that one detection belongs to multiple tracks, or that multiple detections belong to one track.
In this case we interpret the distance to the separating hyperplane as a cost that indicates how well a detection fits to an existing track.
We then use the Hungarian algorithm \cite{Kuhn1955} to find the minimum cost assignment.

\section{Evaluation}
\label{sec:evaluation}

We implemented our approach with Tensorflow.
The implementation of the neural network training script is available as open source software\footnote{\url{https://github.com/fzi-forschungszentrum-informatik/NNAD/tree/tracking/}}.

We train and evaluate our approach on the BDD100K \cite{Yu2020} tracking dataset.
We use the validation set for evaluation because the official test servers do not accept submissions anymore.
This is however valid because we did not perform any hyper-parameter tuning on the validation set.

In theory, our multi-frame model can be trained end-to-end when only using the self-supervised photometric loss to train the optical flow estimation.
We however found that end-to-end training can be unstable.
We therefore employ the following multi-stage training procedure:
\begin{enumerate}
 \item We train the single-frame model.
 The network head for the displacement vector prediction is disabled in this step because the displacement is not observable.
 \item We remove the network head and the second half of the BiFPN, and freeze the remaining parameters.
 Afterwards, we add the optical flow estimation modules, and train them on the synthetic FlyingChairs \cite{Fischer2015} dataset.
 \item We also freeze the optical flow estimation modules, and re-add the second half of the BiFPN and the network heads.
  We train them, including the displacement vector prediction.
\end{enumerate}

We use the following hyper-parameters for each training stage:
\begin{itemize}
 \item Batch size of 8
 \item AdamW optimizer \cite{Kingma2014,Loshchilov2017}
 \item Initial weight decay of $10^{-7}$, scaled proportional with the learning rate
 \item Step-wise learning rate
 \begin{itemize}
  \item $10^{-3}$ until step 600\,000
  \item $10^{-4}$ until step 1\,100\,000
  \item $10^{-5}$ until step 1\,800\,000
 \end{itemize}
 \item Frozen Batch Normalization statistics after step 1\,500\,000
\end{itemize}

\begin{table}
 \begin{center}
 \begin{tabular}{l | r | r}
  \toprule
   & \multicolumn{1}{l |}{AP} & \multicolumn{1}{l}{LAMR} \\
  \midrule
  rider & 0.3665 & 0.5549 \\
  car & 0.8536 & 0.4771 \\
  truck & 0.4503 & 0.6063 \\
  bus & 0.5998 & 0.3996 \\
  motorcycle & 0.3457 & 0.5335 \\
  bicycle & 0.3275 & 0.5188 \\
  \hline
  mean & 0.4906 & 0.5150 \\
  \bottomrule
 \end{tabular}
 \end{center}

 \caption{Object detection results on the BDD100K dataset.
  The metrics are calculated based on an IoU threshold of 0.5.
 }
 \label{tab:det_results}
\end{table}

Table~\ref{tab:det_results} presents the object detection results of our CNN.
It contains both the Average Precision (AP) and the Log-Average Miss Rate (LAMR) \cite{Dollar2012} metrics.
Our CNN achieves good results for the ``car'' class while the other classes achieve a lower average precision.
One reason for this is that cars occur very frequently in the BDD100K dataset while the other classes occur much less often.

We measured the run-time of our model on an \emph{Nvidia Quadro RTX 8000} GPU in mixed-precision mode.
Our single-frame model is fast and processes a single image in 38.7\,ms.
For our multi-frame model, we measured a run-time of 127.9\,ms.
For this measurement, we processed the current frame by the backbone and the first half of the BiFPN, but reused the corresponding feature maps of the previous frame.
The measured run-time of the multi-frame model is much higher than that of the single-frame model.
Most of the additional processing time is however spent in the correlation cost layers.
We use the implementation from Tensorflow Addons which only supports calculations with 32\,bit floating point values.
Notably higher inference speeds should be possible with an optimized implementation at reduced precision.

\begin{table}
 \begin{subtable}{\linewidth}
 \begin{center}
 \begin{tabular}{l | l | r | r | r | r | r}
  \toprule
  \# & metrics  & \multicolumn{1}{l |}{MOTP} & \multicolumn{1}{l |}{MOTA} & \multicolumn{1}{l |}{$r_{\text{m}}$} & \multicolumn{1}{l |}{$r_{\text{fp}}$} & \multicolumn{1}{l}{$r_{\text{mme}}$} \\
  \midrule
  1 & C, KF & 0.207 & 37.0\,\% & 0.459 & 0.144 & 0.026 \\
  2 & C, KF, D & 0.206 & 39.1\,\% & 0.439 & 0.145 & 0.025 \\
  3 & C, KF, E & 0.202 & 36.8\,\% & 0.417 & 0.160 & 0.054 \\
  4 & C, KF, D, E & 0.202 & 36.5\,\% &  0.419 & 0.161 & 0.055 \\
  \bottomrule
 \end{tabular}
 \end{center}
 \caption{MOTA and MOTP values as well as miss ratio $r_{\text{m}}$, false positive ratio $r_{\text{fp}}$, and mismatch ratio $r_{\text{mme}}$.}
 \end{subtable}

 \vspace*{0.5cm}
 
 \begin{subtable}{\linewidth}
 \begin{center}
 \begin{tabular}{l | l | r | r | r}
  \toprule
  \# & metrics & \multicolumn{1}{l |}{mostly tracked} & \multicolumn{1}{l |}{partially tracked} & \multicolumn{1}{l}{mostly lost} \\
  \midrule
  1 & C, KF & 14.5\,\% & 45.5\,\% & 39.9\,\% \\
  2 & C, KF, D & 16.9\,\% & 45.9\,\% & 37.1\,\% \\
  3 & C, KF, E & 22.2\,\% & 46.2\,\% & 31.5\,\% \\
  4 & C, KF, D, E & 21.9\,\% & 46.3\,\% & 31.8\,\% \\
  \bottomrule
 \end{tabular}
 \end{center}
 \caption{Percentage of successfully tracked objects.
 Here, an object is mostly tracked if it is tracked for at least 80\,\% of its ground truth lifespan.
 Analogously, it is mostly lost if it is tracked for less than 20\,\% of its lifespan.
 Otherwise, the object is partially tracked.
 }
 \end{subtable}

 \caption{Object tracking results over all classes on the BDD100K tracking dataset.
 We perform several experiments to evaluate the contributions of our different assignment metrics.
 We use the following abbreviations for the assignment metrics in these tables:
 C: Use class cost.
 KF: Use the Mahalanobis distance between the Kalman filter state and the new detection.
 D: Use the displacement vector based distance.
 E: Use the embedding vector distance that describes object similarity.
 }
 \label{tab:tracking_results}
\end{table}

The results of our tracking approach when using different assignment metrics can be found in Table~\ref{tab:tracking_results}.
We report the commonly used Multiple Object Tracking Accuracy (MOTA) and Multiple Object Tracking Precision (MOTP) metrics \cite{Bernardin2008}.
Please note that we follow the original definition of MOTP while some other publications report $100 \cdot (1 - \text{MOTP})$.
We also report the miss ratio, the false positive ratio, and the mismatch ratio which are used to calculate the MOTA metric, as well as the percentage of successfully tracked objects.

The first experiment establishes the baseline.
It uses only the class costs and the Mahalanobis distance between the Kalman filter states and the new detections during the assignment step.
This information is available with any object detector, while the other experiments require the additional metrics that our models predict.

The second experiment adds the displacement vector based distance to the assignment step.
The additional information allows to make more informed associations.
This reduces the miss ratio and increases the MOTA value.
Also the percentage of mostly tracked objects increases.

The third experiment uses the class distance, the filter state, and the embedding vector during the assignment step.
Many more detections can be assigned to existing tracks by not only trusting in geometric constraints, but also in visual appearance features.
This reduces the miss rate significantly compared to both the baseline and experiment 2.
It results in longer tracks, and a higher percentage of mostly tracked and partially tracked objects.
It also reduces the localization error as given by the MOTP metric.
This is because the more frequent successful associations lead to more frequent updates of the Kalman filter.
But trusting the visual appearance features also results in more identity switches, and therefore in a lower MOTA value.

The fourth experiment combines all metrics in the assignment step.
Interestingly, it does not perform better than experiment 3.
The displacement vector based metric does not provide additional information in presence of the embedding vector based metric.
One explanation for this is that both the embedding vector and the displacement vector are calculated based on the same visual input.
They might therefore have similar failure cases and can be correlated.

Based on these experiments, we recommend to use the combination of metrics from experiment 2 if a high MOTA value is desired.
The combination of metrics from experiment 3 should be chosen if a low localization error, a higher recall, and longer tracks are desired.
This also allows to use our fast single-frame model.
Combining all metrics does not improve the results in our experiments, but could still be beneficial when used with a different CNN architecture.

\section{Conclusions}
\label{sec:conclusion}

We presented a simple, fast, and accurate tracking-by-detection approach that relies on visual and motion cues.
Our single-frame model predicts embedding vectors that allow to associate objects based on their visual appearance.
It requires little computation time and reduces miss rate and localization error significantly compared to the baseline.
Our multi-frame model predicts displacement vectors between consecutive video frames for each object detection.
This also provides important information to assign new detections to existing tracks.
Compared to the baseline, it reduces the miss rate and improves the MOTA metric significantly.
We successfully employ our approach inside the vision stack of a self-driving research vehicle.

\section*{Acknowledgment}
The author conducted parts of this research during his stay at LISA $\times$ CVRR, University of California San Diego (UCSD).
His stay abroad was financially supported by the Karlsruhe House of Young Scientists (KHYS).

\bibliographystyle{IEEEtran}
\bibliography{content/mybibliography}

\begin{thebibliography}{10}
\providecommand{\url}[1]{#1}
\csname url@samestyle\endcsname
\providecommand{\newblock}{\relax}
\providecommand{\bibinfo}[2]{#2}
\providecommand{\BIBentrySTDinterwordspacing}{\spaceskip=0pt\relax}
\providecommand{\BIBentryALTinterwordstretchfactor}{4}
\providecommand{\BIBentryALTinterwordspacing}{\spaceskip=\fontdimen2\font plus
\BIBentryALTinterwordstretchfactor\fontdimen3\font minus
  \fontdimen4\font\relax}
\providecommand{\BIBforeignlanguage}[2]{{%
\expandafter\ifx\csname l@#1\endcsname\relax
\typeout{** WARNING: IEEEtran.bst: No hyphenation pattern has been}%
\typeout{** loaded for the language `#1'. Using the pattern for}%
\typeout{** the default language instead.}%
\else
\language=\csname l@#1\endcsname
\fi
#2}}
\providecommand{\BIBdecl}{\relax}
\BIBdecl

\bibitem{Bolme2010}
D.~S. Bolme, J.~R. Beveridge, B.~A. Draper, and Y.~M. Lui, ``Visual object
  tracking using adaptive correlation filters,'' in \emph{The Twenty-Third
  {IEEE} Conference on Computer Vision and Pattern Recognition (CVPR)
  2010}.\hskip 1em plus 0.5em minus 0.4em\relax {IEEE} Computer Society, 2010,
  pp. 2544--2550.

\bibitem{Henriques2015}
J.~F. Henriques, R.~Caseiro, P.~Martins, and J.~Batista, ``High-speed tracking
  with kernelized correlation filters,'' \emph{IEEE Transactions on Pattern
  Analysis and Machine Intelligence}, vol.~37, no.~3, pp. 583--596, 2015.

\bibitem{Danelljan2015}
M.~Danelljan, G.~H{\"{a}}ger, F.~S. Khan, and M.~Felsberg, ``Convolutional
  features for correlation filter based visual tracking,'' in \emph{2015 {IEEE}
  International Conference on Computer Vision (ICCV) Workshop}.\hskip 1em plus
  0.5em minus 0.4em\relax {IEEE} Computer Society, 2015, pp. 621--629.

\bibitem{Bertinetto2016}
L.~Bertinetto, J.~Valmadre, J.~F. Henriques, A.~Vedaldi, and P.~H.~S. Torr,
  ``Fully-convolutional siamese networks for object tracking,'' in
  \emph{Computer Vision - {ECCV} 2016 Workshops}, G.~Hua and H.~J{\'{e}}gou,
  Eds.\hskip 1em plus 0.5em minus 0.4em\relax Springer International
  Publishing, 2016, pp. 850--865.

\bibitem{Dong2018}
X.~Dong and J.~Shen, ``Triplet loss in siamese network for object tracking,''
  in \emph{Computer Vision - {ECCV} 2018 - 15th European Conference},
  V.~Ferrari, M.~Hebert, C.~Sminchisescu, and Y.~Weiss, Eds.\hskip 1em plus
  0.5em minus 0.4em\relax Springer, 2018, pp. 472--488.

\bibitem{He2018}
A.~He, C.~Luo, X.~Tian, and W.~Zeng, ``A twofold siamese network for real-time
  object tracking,'' in \emph{2018 {IEEE} Conference on Computer Vision and
  Pattern Recognition (CVPR)}.\hskip 1em plus 0.5em minus 0.4em\relax {IEEE}
  Computer Society, 2018, pp. 4834--4843.

\bibitem{Li2018}
B.~Li, J.~Yan, W.~Wu, Z.~Zhu, and X.~Hu, ``High performance visual tracking
  with siamese region proposal network,'' in \emph{2018 {IEEE} Conference on
  Computer Vision and Pattern Recognition (CVPR)}.\hskip 1em plus 0.5em minus
  0.4em\relax {IEEE} Computer Society, 2018, pp. 8971--8980.

\bibitem{Zhang2008}
L.~Zhang, Y.~Li, and R.~Nevatia, ``Global data association for multi-object
  tracking using network flows,'' in \emph{2008 {IEEE} Computer Society
  Conference on Computer Vision and Pattern Recognition (CVPR)}.\hskip 1em plus
  0.5em minus 0.4em\relax {IEEE} Computer Society, 2008, pp. 1--8.

\bibitem{Pirsiavash2011}
H.~Pirsiavash, D.~Ramanan, and C.~C. Fowlkes, ``Globally-optimal greedy
  algorithms for tracking a variable number of objects,'' in \emph{The 24th
  {IEEE} Conference on Computer Vision and Pattern Recognition (CVPR)
  2011}.\hskip 1em plus 0.5em minus 0.4em\relax {IEEE} Computer Society, 2011,
  pp. 1201--1208.

\bibitem{Butt2013}
A.~A. Butt and R.~T. Collins, ``Multi-target tracking by lagrangian relaxation
  to min-cost network flow,'' in \emph{2013 {IEEE} Conference on Computer
  Vision and Pattern Recognition (CVPR)}.\hskip 1em plus 0.5em minus
  0.4em\relax {IEEE} Computer Society, 2013, pp. 1846--1853.

\bibitem{Kalman1960}
R.~E. Kalman, ``A new approach to linear filtering and prediction problems,''
  \emph{Journal of Basic Engineering}, vol.~82, no.~1, pp. 35--45, 1960.

\bibitem{Bewley2016}
A.~Bewley, Z.~Ge, L.~Ott, F.~T. Ramos, and B.~Upcroft, ``Simple online and
  realtime tracking,'' in \emph{2016 {IEEE} International Conference on Image
  Processing (ICIP)}.\hskip 1em plus 0.5em minus 0.4em\relax IEEE, 2016, pp.
  3464--3468.

\bibitem{Wojke2017}
N.~Wojke, A.~Bewley, and D.~Paulus, ``Simple online and realtime tracking with
  a deep association metric,'' in \emph{2017 {IEEE} International Conference on
  Image Processing (ICIP)}.\hskip 1em plus 0.5em minus 0.4em\relax IEEE, 2017,
  pp. 3645--3649.

\bibitem{Feichtenhofer2017}
C.~Feichtenhofer, A.~Pinz, and A.~Zisserman, ``Detect to track and track to
  detect,'' in \emph{2017 IEEE International Conference on Computer Vision
  (ICCV)}.\hskip 1em plus 0.5em minus 0.4em\relax IEEE Computer Society, 2017,
  pp. 3057--3065.

\bibitem{Bergmann2019}
P.~Bergmann, T.~Meinhardt, and L.~Leal{-}Taix{\'{e}}, ``Tracking without bells
  and whistles,'' in \emph{2019 {IEEE/CVF} International Conference on Computer
  Vision (ICCV)}, {IEEE}, Ed., 2019, pp. 941--951.

\bibitem{Luo2018}
W.~Luo, B.~Yang, and R.~Urtasun, ``Fast and furious: Real time end-to-end 3d
  detection, tracking and motion forecasting with a single convolutional net,''
  in \emph{2018 {IEEE} Conference on Computer Vision and Pattern Recognition
  (CVPR)}.\hskip 1em plus 0.5em minus 0.4em\relax {IEEE} Computer Society,
  2018, pp. 3569--3577.

\bibitem{Tan2020}
M.~Tan, R.~Pang, and Q.~V. Le, ``Efficientdet: Scalable and efficient object
  detection,'' in \emph{2020 {IEEE/CVF} Conference on Computer Vision and
  Pattern Recognition (CVPR)}.\hskip 1em plus 0.5em minus 0.4em\relax {IEEE}
  Computer Society, 2020, pp. 10\,778--10\,787.

\bibitem{Tan2019}
M.~Tan and Q.~V. Le, ``Efficientnet: Rethinking model scaling for convolutional
  neural networks,'' in \emph{Proceedings of the 36th International Conference
  on Machine Learning (ICML)}, K.~Chaudhuri and R.~Salakhutdinov, Eds.\hskip
  1em plus 0.5em minus 0.4em\relax PMLR, 2019, pp. 6105--6114.

\bibitem{Lin2017_2}
T.~Lin, P.~Doll{\'{a}}r, R.~B. Girshick, K.~He, B.~Hariharan, and S.~J.
  Belongie, ``Feature pyramid networks for object detection,'' in \emph{2017
  {IEEE} Conference on Computer Vision and Pattern Recognition (CVPR)}.\hskip
  1em plus 0.5em minus 0.4em\relax {IEEE} Computer Society, 2017, pp. 936--944.

\bibitem{Liu2018}
S.~Liu, L.~Qi, H.~Qin, J.~Shi, and J.~Jia, ``Path aggregation network for
  instance segmentation,'' in \emph{2018 {IEEE} Conference on Computer Vision
  and Pattern Recognition (CVPR)}.\hskip 1em plus 0.5em minus 0.4em\relax
  {IEEE} Computer Society, 2018, pp. 8759--8768.

\bibitem{Ghiasi2019}
G.~Ghiasi, T.~Lin, and Q.~V. Le, ``{NAS-FPN:} learning scalable feature pyramid
  architecture for object detection,'' in \emph{2019 {IEEE} Conference on
  Computer Vision and Pattern Recognition (CVPR)}.\hskip 1em plus 0.5em minus
  0.4em\relax Computer Vision Foundation / {IEEE}, 2019, pp. 7036--7045.

\bibitem{Lin2017}
T.~Lin, P.~Goyal, R.~B. Girshick, K.~He, and P.~Doll{\'{a}}r, ``Focal loss for
  dense object detection,'' in \emph{{IEEE} International Conference on
  Computer Vision (ICCV) 2017}.\hskip 1em plus 0.5em minus 0.4em\relax {IEEE}
  Computer Society, 2017, pp. 2999--3007.

\bibitem{Wu2017}
R.~Manmatha, C.~Wu, A.~J. Smola, and P.~Kr{\"{a}}henb{\"{u}}hl, ``Sampling
  matters in deep embedding learning,'' in \emph{{IEEE} International
  Conference on Computer Vision (ICCV) 2017}.\hskip 1em plus 0.5em minus
  0.4em\relax {IEEE} Computer Society, 2017, pp. 2859--2867.

\bibitem{Salscheider2020_2}
\BIBentryALTinterwordspacing
N.~O. Salscheider, ``Featurenms: Non-maximum suppression by learning feature
  embeddings,'' 2020. [Online]. Available:
  \url{https://arxiv.org/abs/2002.07662}
\BIBentrySTDinterwordspacing

\bibitem{Kendall2017}
A.~Kendall, Y.~Gal, and R.~Cipolla, ``Multi-task learning using uncertainty to
  weigh losses for scene geometry and semantics,'' in \emph{2018 {IEEE}
  Conference on Computer Vision and Pattern Recognition (CVPR)}.\hskip 1em plus
  0.5em minus 0.4em\relax {IEEE} Computer Society, 2018, pp. 7482--7491.

\bibitem{Sun2018}
D.~Sun, X.~Yang, M.~Liu, and J.~Kautz, ``Pwc-net: Cnns for optical flow using
  pyramid, warping, and cost volume,'' in \emph{2018 {IEEE} Conference on
  Computer Vision and Pattern Recognition (CVPR)}.\hskip 1em plus 0.5em minus
  0.4em\relax {IEEE} Computer Society, 2018, pp. 8934--8943.

\bibitem{Huang2017}
G.~Huang, Z.~Liu, L.~van~der Maaten, and K.~Q. Weinberger, ``Densely connected
  convolutional networks,'' in \emph{2017 {IEEE} Conference on Computer Vision
  and Pattern Recognition (CVPR)}.\hskip 1em plus 0.5em minus 0.4em\relax
  {IEEE} Computer Society, 2017, pp. 2261--2269.

\bibitem{Godard2019}
\BIBentryALTinterwordspacing
C.~Godard, O.~M. Aodha, M.~Firman, and G.~Brostow, ``Digging into
  self-supervised monocular depth estimation,'' 2019. [Online]. Available:
  \url{https://arxiv.org/abs/1806.01260}
\BIBentrySTDinterwordspacing

\bibitem{Abbeel2005}
P.~Abbeel, A.~Coates, M.~Montemerlo, A.~Y. Ng, and S.~Thrun, ``Discriminative
  training of kalman filters,'' in \emph{Robotics: Science and Systems I},
  S.~Thrun, G.~S. Sukhatme, and S.~Schaal, Eds.\hskip 1em plus 0.5em minus
  0.4em\relax The {MIT} Press, 2005, pp. 289--296.

\bibitem{Kuhn1955}
H.~W. Kuhn, ``The hungarian method for the assignment problem,'' \emph{Naval
  Research Logistics Quarterly}, vol.~2, no. 1-2, pp. 83--97, 1955.

\bibitem{Yu2020}
F.~Yu, H.~Chen, X.~Wang, W.~Xian, Y.~Chen, F.~Liu, V.~Madhavan, and T.~Darrell,
  ``{BDD100K:} {A} diverse driving dataset for heterogeneous multitask
  learning,'' in \emph{2020 {IEEE/CVF} Conference on Computer Vision and
  Pattern Recognition (CVPR)}.\hskip 1em plus 0.5em minus 0.4em\relax {IEEE}
  Computer Society, 2020, pp. 2633--2642.

\bibitem{Fischer2015}
A.~Dosovitskiy, P.~Fischer, E.~Ilg, P.~H{\"{a}}usser, C.~Hazirbas, V.~Golkov,
  P.~van~der Smagt, D.~Cremers, and T.~Brox, ``Flownet: Learning optical flow
  with convolutional networks,'' in \emph{2015 {IEEE} International Conference
  on Computer Vision (ICCV)}.\hskip 1em plus 0.5em minus 0.4em\relax {IEEE}
  Computer Society, 2015, pp. 2758--2766.

\bibitem{Kingma2014}
\BIBentryALTinterwordspacing
D.~P. Kingma and J.~Ba, ``Adam: {A} method for stochastic optimization,'' 2015.
  [Online]. Available: \url{https://arxiv.org/abs/1412.6980}
\BIBentrySTDinterwordspacing

\bibitem{Loshchilov2017}
\BIBentryALTinterwordspacing
I.~Loshchilov and F.~Hutter, ``Decoupled weight decay regularization,'' 2019.
  [Online]. Available: \url{https://arxiv.org/abs/1711.05101}
\BIBentrySTDinterwordspacing

\bibitem{Dollar2012}
P.~Doll{\'{a}}r, C.~Wojek, B.~Schiele, and P.~Perona, ``Pedestrian detection:
  An evaluation of the state of the art,'' \emph{IEEE Transactions on Pattern
  Analysis and Machine Intelligence}, vol.~34, no.~4, pp. 743--761, 2017.

\bibitem{Bernardin2008}
K.~Bernardin and R.~Stiefelhagen, ``Evaluating multiple object tracking
  performance: The {CLEAR} {MOT} metrics,'' \emph{EURASIP Journal on Image and
  Video Processing}, pp. 1--10, 2008.

\end{thebibliography}

\end{document}